\documentclass[journal]{IEEEtran}
\ifCLASSINFOpdf
\else
\fi
  \usepackage[caption=false,font=footnotesize]{subfig}
\usepackage{color}
\usepackage{textcomp}
\usepackage{amssymb}
\usepackage{versions}
\usepackage{longtable}
\usepackage{booktabs}
\usepackage{mathtools}
    \usepackage{mathtools}        
    \usepackage{%
       multirow,                   
       tabu                        
    }

    \usepackage{graphicx}
\newcommand{\plotbar}[1]{\textcolor{plotcol#1}{\rule[2pt]{12pt}{2pt}}}

\definecolor{mlcolor1}{rgb}{0.8500,0.3250,0.0980}
\definecolor{mlcolor2}{rgb}{0.9290,0.6940,0.1250}
\definecolor{mlcolor3}{rgb}{0.4940,0.1840,0.5560}
\definecolor{mlcolor4}{rgb}{0.96863,0.83137,0.77647}

\definecolor{plotcol1}{rgb}{0.84314,0.098039,0.1098}
\definecolor{plotcol2}{rgb}{0.99216,0.68235,0.38039}
\definecolor{plotcol3}{rgb}{1,1,0.74902}
\definecolor{plotcol4}{rgb}{0.67059,0.86667,0.64314}
\definecolor{plotcol5}{rgb}{0.16863,0.51373,0.72941}

\DeclarePairedDelimiter\abs{\lvert}{\rvert}%
\DeclarePairedDelimiterX{\norm}[1]{\lVert}{\rVert}{#1}
\excludeversion{draft}
\newcommand{\igdp}{IGD$^+$}
\usepackage{hyperref}
\usepackage{cleveref}

\renewcommand{\textrightarrow}{$\rightarrow$}
\begin{document}
%
\title{Scalable Many-Objective Pathfinding Benchmark Suite}
%
%
%

\author{Jens~Weise,~\IEEEmembership{Member,~IEEE},
        and~Sanaz~Mostaghim,~\IEEEmembership{Senior~Member,~IEEE}
\thanks{J. Weise and S. Mostaghim are with the Institute for Intelligent Cooperating Systems, Otto von Guericke University Magdeburg, Germany e-mail: \{jens.weise,sanaz.mostaghim\}@ovgu.de.}%
}

\maketitle

\begin{abstract}
THIS IS A PREPRINT SUBMITTED TO arxiv.org. IT CAN BE SUBSTITUTED WITH A NEWER VERSION AT ANY TIME.\\
Route planning also known as pathfinding is one of the key elements in logistics, mobile robotics and other applications, where engineers face many conflicting objectives. However, most of the current route planning algorithms consider only up to three objectives. In this paper, we propose a scalable many-objective benchmark problem covering most of the important features for routing applications based on real-world data. We define five objective functions representing distance, traveling time, delays caused by accidents, and two route specific features such as curvature and elevation. We analyse several different instances for this test problem and provide their true Pareto-front to analyse the problem difficulties. We apply three well-known evolutionary multi-objective algorithms. Since this test benchmark can be easily transferred to real-world routing problems, we construct a routing problem from OpenStreetMap data. We evaluate the three optimisation algorithms and observe that we are able to provide promising results for such a real-world application. The proposed benchmark represents a scalable many-objective route planning optimisation problem enabling researchers and engineers to evaluate their many-objective approaches.
\end{abstract}


%
\IEEEpeerreviewmaketitle

\section{Introduction}
Optimal route planning (or pathfinding) is among the most challenging tasks for industrial and logistical applications~\cite{Toth2014VehicleApplications}. Any improvement in the results can have a considerable impact on many factors, such as fuel consumption and the environment. The current state-of-the-art route planning algorithms usually consider the travel time and the distance in the optimisation.
However, specific applications encounter additional criteria such as the curvature of the route, the elevation (ascent), or environmental issues such as air pollution caused by fuel consumption. These criteria can profoundly influence the practicability of the solutions. For instance, for animal transportation, we need to additionally minimise the number of curves in the route (or maximise the smoothness). 
Reducing the length of the route can help to reduce fuel consumption, while possibly increasing the travelling time. Other criteria such as the ascent of a path can be considered for heavy vehicles which can consume more fuel on such non-flat routes. 

The goal of this paper is to propose a many-objective route planing problem representing five objective functions which at the same time are highly related to their real-world counterparts. This real-world problem can be considered as scalable in terms of complexity, since the size of the search space can be varied, influencing the objective functions as well in terms of search space size.
To the best of our knowledge, there is no work in the literature which considers all of these criteria at the same time. 
Similar to the existing navigation and route planning algorithms, we work on a graph-based approach for addressing this many-objective problem.  
We additionally apply the benchmark characteristics to the real-world data from OpenStreetMap in Berlin.\\ 
Our results show that this problem can be used both as a benchmark and as well as a real-world application. We additionally provide the true Pareto-front of 491 benchmark instances, their respective Pareto-sets and the code to generate specific instances of the proposed benchmark.

The paper is structured as follows. 
In Section \ref{sec:rw}, we provide an overview of the related works. \Cref{sec:main} is dedicated to the many-objective pathfinding problem, the proposed encoding and the objective functions. In \cref{sec:experiments}, we provide experiments using three state-of-the-art optimisation algorithms, and in Section \ref{sec:realworld}, we transfer the benchmark and objective functions to real-world road map data. \Cref{sec:conclusion} concludes the paper and gives an overview of future work.



\section{Related Works}
\label{sec:rw}
There is an extensive amount of literature in the field of route planning and pathfinding in general and especially for vehicle route planning which uses evolutionary algorithms~\cite{Ahmed2013Multi-objectiveAlgorithms,Castillo2007MultipleRobots,Alajlan2013GlobalAlgorithms, Ahmed2011Multi-objectiveRepresentation}. 
The most important feature concerns the solution representation, which can define the size of the search space and influence the efficiency of the algorithms. 

Various representations such as graph-based \cite{Changan2010DynamicAlgorithm,Qiongbing2016AProblems,Tozer2017Many-objectiveLearning,Pulido2015DimensionalitySearch,Rajabi-Bahaabadi2015Multi-objectiveAlgorithm,Weise2018}, 
and grid-based representations ~\cite{Ahmed2013Multi-objectiveAlgorithms,Yakovlev2015Grid-BasedPlanning} have been suggested for the pathfinding problem. 
Typically, there are two main approaches: The first is a variable-length chromosome representation which is often used in combination with the graph-based problem representation~\cite{Elshamli2004GeneticPlanning,Jun2010Multi-objectiveAlgorithm,ShashiMittal2007Three-dimensionalAlgorithms,Davoodi2013Multi-objectiveSpace}. This approach represents a solution as a list of nodes, which can be of different length when computing a path. The second approach is a fixed-length chromosome, representing the  directions of travel together with a list of nodes in a graph or a list of grid cells~\cite{Besada-Portas2013OnPlanners,Qu2013AnRobots, Ahmed2013Multi-objectiveAlgorithms}.
Grid-based representations for pathfinding problems are shown to be very practical for evolutionary algorithms~\cite{Ahmed2013Multi-objectiveAlgorithms,Ahmed2011Multi-objectiveRepresentation}. Such grid representations can be refined depending on the required resolution of the problem. Moreover, they are often used for benchmarking purposes~\cite{Sturtevant2012BenchmarksPathfinding,Koceski2014AFinding}. Also, they can represent the real-world problems abstractly by discretising the problem representation~\cite{Anguelov2011VideoMaps}. 
Grids typically consist of units with adjustable sizes~\cite{R2013PathGraph}. An encoding can consist of a linked-list of units~\cite{Xiao1999AnNavigation}, the directions~\cite{Ahmed2013Multi-objectiveAlgorithms}, or the coordinates of several waypoints.\\
It is comparatively easy to convert a grid into a graph by considering units as nodes and their contact-edges as the graph's edges. This is done in several applications, e.g. the game industry when it comes to pathfinding, by superimposing a grid over an area and using graph-search algorithms~\cite{Yap2002Grid-BasedPath-Finding}. The commonly used A$^*$ algorithm is an example of pathfinding on a grid which is transferred to a graph~\cite{Yap2002Grid-BasedPath-Finding,Sturtevant2012BenchmarksPathfinding}.\\
In general, graph-based representations allow a higher flexibility in representing real-world problems, which can be considered heterogeneous, compared to grids which are usually homogeneous. Due to this, in this paper, we present the proposed benchmark problem as a grid transferred to a graph and facilitate the methods to evaluate them on realistic graph-represented road map data.

Route planning techniques can be applied to applications as pipe- or wire-routing~\cite{Belov2019PositionRouting}. Weise et al. used a customised NSGA-II algorithm for a multi-objective generation of wiring harnesses \cite{Weise2019} and optimised length and the maximum heat for a wire which is confronted in a path. Oleiwi et al. used a hybrid approach by modifying a genetic algorithm based on the A* algorithm \cite{Oleiwi2014ModifiedPlanning}. Changan and Quiongbing used genetic algorithms and customised operators to work on the pathfinding problem~\cite{Changan2010DynamicAlgorithm,Qiongbing2016AProblems}.\\
Considering many-objective pathfinding problems, there is a limited amount of literature using evolutionary algorithms. Tozer et al.~\cite{Tozer2017Many-objectiveLearning} provide an overview of existing approaches and use reinforcement learning to address the problem with six objective functions. Pulido et al.~\cite{Pulido2015DimensionalitySearch} introduce a dimensionality reduction technique in order to minimise dominance checks during the optimisation and tested their algorithm on a map from a real-world application. They extended the NAMOA$^*$ algorithm, which was first introduced by Mandow~\cite{Mandow2010MultiobjectiveHeuristics} and is a multi-objective extension to the well-known A$^*$ algorithm~\cite{Hart1968APaths}. Considering many-objective benchmarks in general, there are several existing benchmark frameworks and functions available~\cite{Fieldsend2019AGenerator, DebScalableOptimization}.

\section{Many-Objective Pathfinding Problem}
\label{sec:main}
In this section, we propose a many-objective pathfinding problem which can be additionally used as a benchmark problem with a scalable size of the search space. As the benchmark aims to represent environments for pathfinding algorithms for maps, we construct the instances by defining a cartesian grid with a specific size, where each cell has the same dimensions, also known as \emph{integer lattice}. The variable properties of the benchmark influence the properties of each cell in the lattice.

\subsection{Benchmark problem construction}
The multi-objective route planning problem, hereafter called \textit{pathfinding problem}, can be defined as a network-flow problem~\cite{Raith2009AProblems, Pulido2015DimensionalitySearch}. The goal is to find a set of optimal paths (routes) $P*=\{p_1,\cdots,p_L\}$ in a graph $G(V,E)$ from a starting node $n_S \in V$ to a pre-defined end node $n_{End} \in V$, i.e.,  $p_i=(n_S,\cdots,n_{End})$ for path $N_i$. 
Before constructing the problem-related graph, we model a grid which is used as a map for the pathfinding problem. 
We assume to have a two-dimensional search space defined by a given size (i.e. size of the map) denoted by the range $[x_{min}, y_{min}]$ and $[x_{max}, y_{max}]$, $x,y\in\mathbb{N}$. This search space is divided into grid cells which define the resolution of the path planning and therefore, the size of the search space.   
We define different types of grid cells which impose constraints on the velocity of movements indicated by $v_{max}$ representing different road types as well as obstacles where a movement cannot occur.

We furthermore define an elevation function $h$ with a variable number of hills which can be defined by either using a peak-function or a combination of hill functions. 
Two more features concern the neighborhood and backtracking. 
The neighbourhood property defines the possible neighbour cells to which an agent can move. We use the $2^k$-neighbourhood similar to \cite{Stern2019Multi-AgentBenchmarksb}. In this case, $k=2$ means that it is possible to go to one of the four neighbours, located in the cardinal directions, where $k= 3$ defines eight possible neighbours, taking the diagonal cells into account.
The backtracking property of the benchmark defines if an agent can go backwards or only forward. For instance, if backtracking is allowed and the goal is to go from the north-west corner of the grid to the southern-east one, the agent can go in any direction specified by the $2^k$-neighourhood from any cell on a certain path. If backtracking is not allowed, the agent can only move in the directions of east, south and south-east (if $k=3$). An 8-neighbourhood with enabled backtracking is also known as \emph{king-moves}, derived from chess.

\begin{figure}[t]
     \centering
\subfloat[Grid graph of a K2 instance (known as \emph{rook's graph})]
{
\label{fig:grid:grid}
\includegraphics[width=0.2\textwidth]{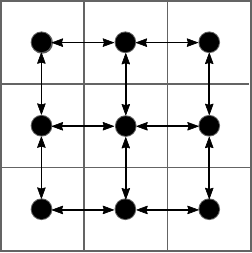}
}\quad
\subfloat[Grid graph of a K3 instance (known as \emph{king's graph})]
{
\label{fig:grid:graph}
\includegraphics[width=0.2\textwidth]{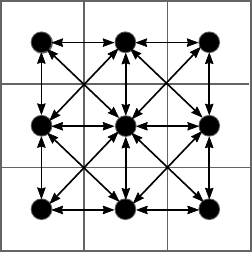}
}%
\caption{Superimposed graphs on their respective grids}
\label{fig:grid}
\end{figure}

In the following, we propose a graph-based representation of the benchmark grid. Therefore, we describe all objectives for the evaluation of a \textit{solution} represented as a path $N$ of length $K$ consisting of a list of adjacent nodes in a graph $G=(V,E)$: $N=(n_i,n_{i+1},\cdots,n_k)$. However, for the evaluation on the grid (as described above), the nodes $n_i$ can be replaced by their respective coordinates $(x_i, y_i)$. Each cell $i$ of the grid is represented as a node $n_i$ located on the coordinate $(x_i, y_i)$, and depending on the $2^k$-neighbourhood and backtracking property, the corresponding neighbours are connected using edges. The resulting graph is also known as \emph{grid graph} or \emph{lattice graph}. \Cref{fig:grid} visually shows an example of the transfer from a grid to a graph.

\begin{figure*}[t]
     \centering
\subfloat[three different types of cells ($v_{max} (blue) > v_{max} (turquoise) > v_{max} (bright yellow) > 0$.]
{
\label{fig:map:streets}
\includegraphics[width=0.3\textwidth]{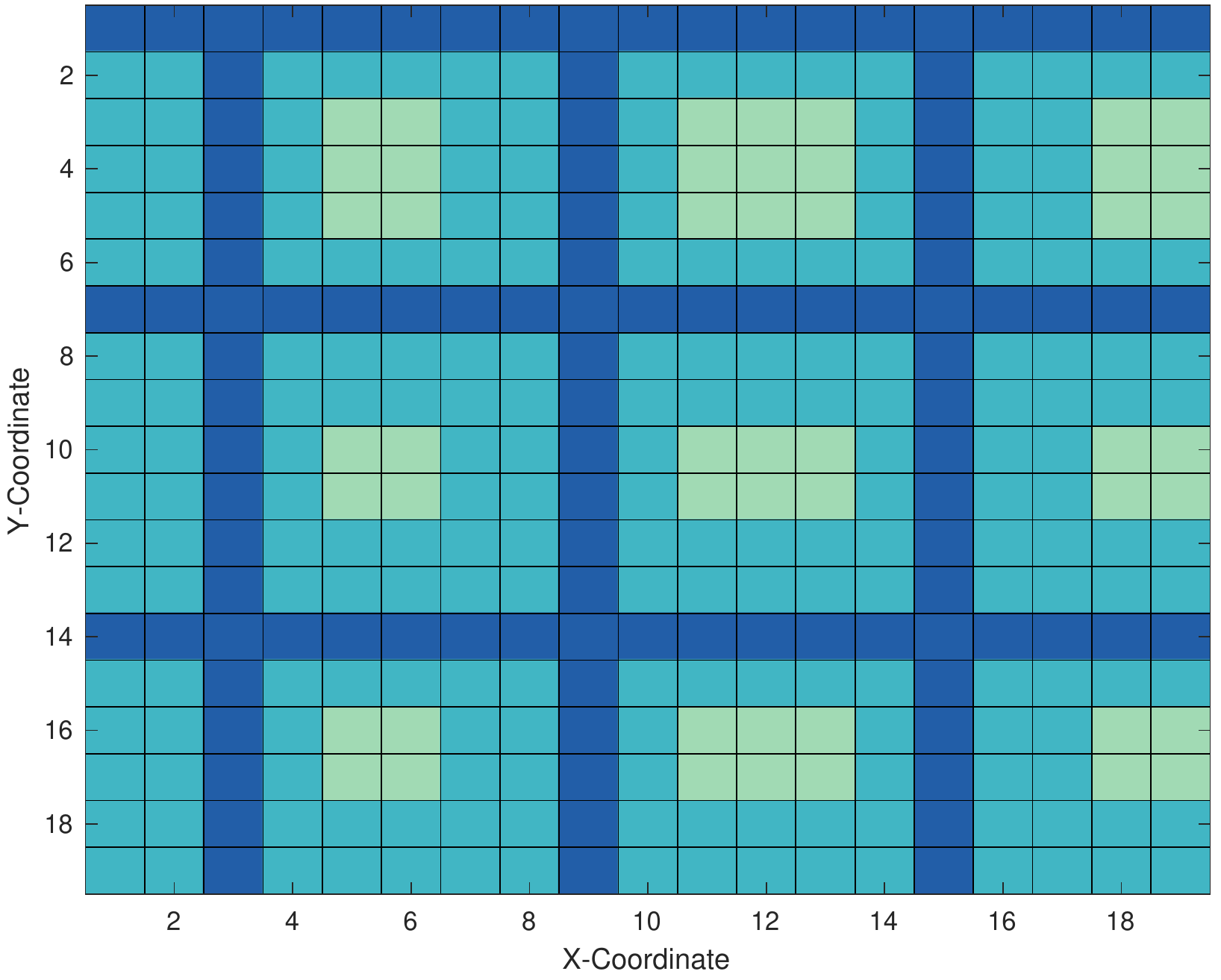}
}\quad
\subfloat[\emph{lake}-obstacle with $v_{max} = 0$ in the middle of the map.]
{
\label{fig:map:lake}
\includegraphics[width=0.3\textwidth]{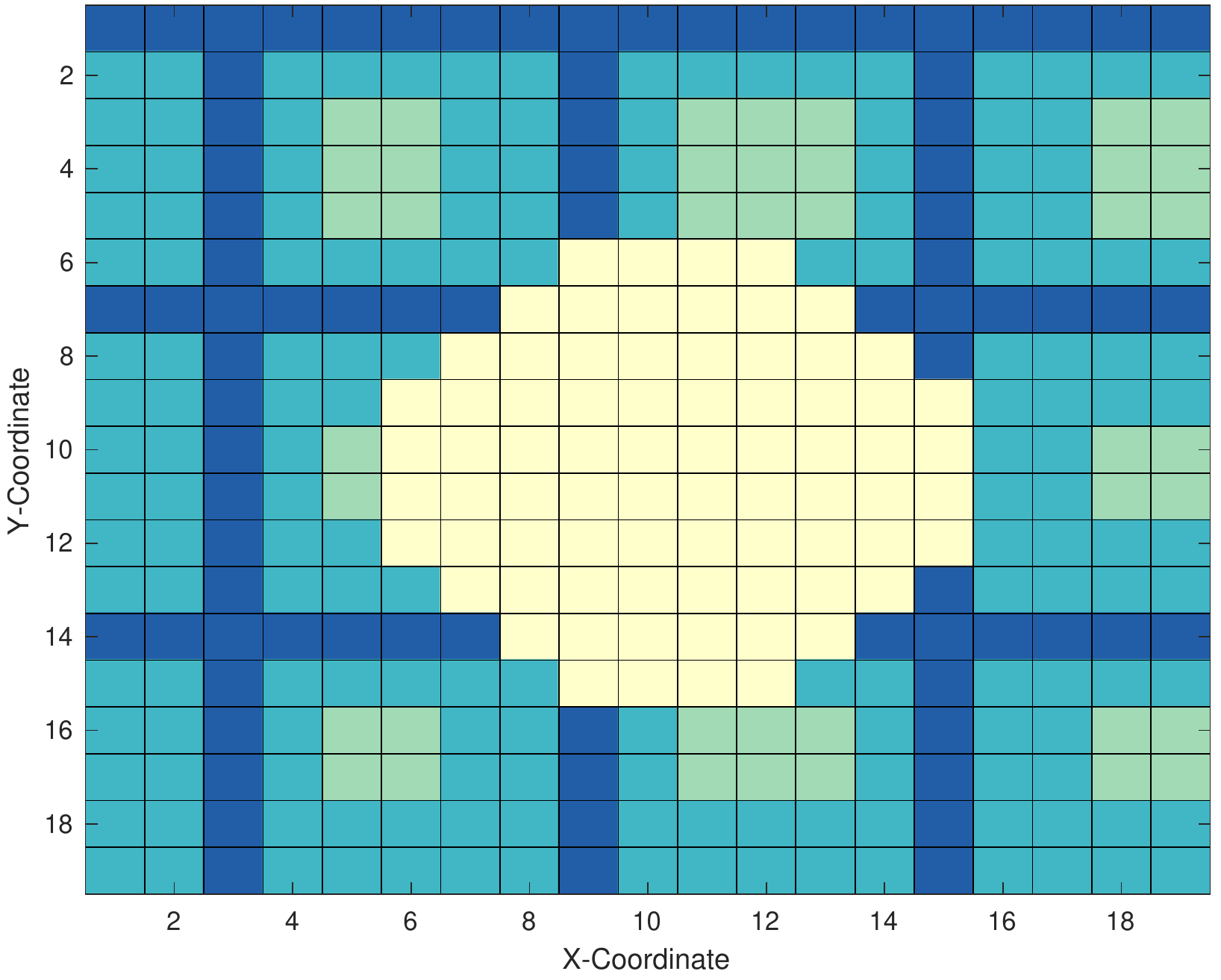}
}\quad
\subfloat[\emph{checkboard}-obstacles representing a block-like infrastructure]
{
\label{fig:map:checkboard}
\includegraphics[width=0.3\textwidth]{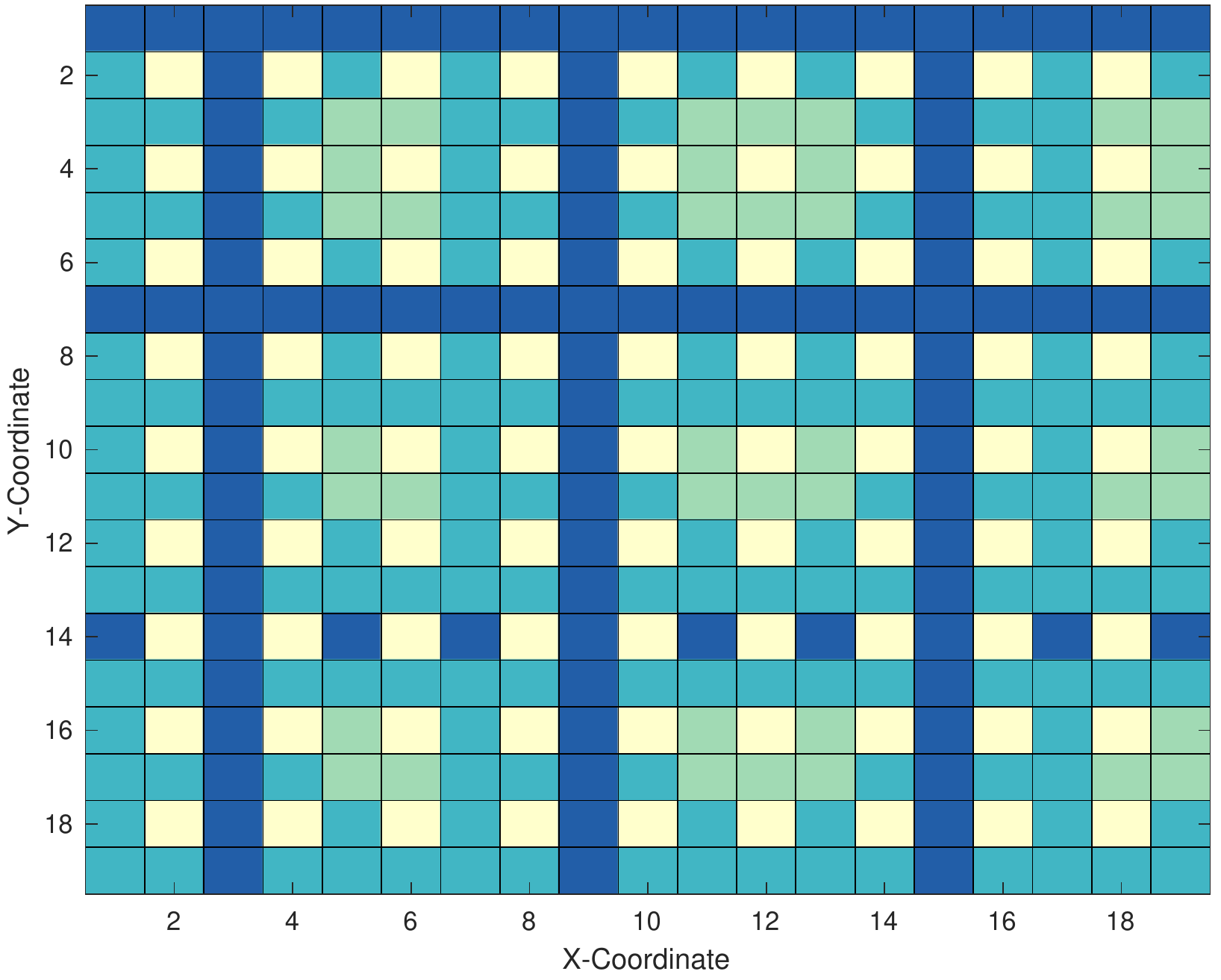}
}%
\caption{Different grid cell properties}\label{fig:maps}
\end{figure*}

\subsection{Objective functions}
\label{sec:objectives}
Given a solution represented by a path $N$ of length $K$ as $N=(n_i,n_{i+1},\cdots,n_k)$, we can evaluate it by five objectives to be minimised: (1) Euclidean length, (2) Delays, (3) Elevation, (4) Traveling time and (5) Smoothness (Curvature). 

\par
\noindent
\textbf{Objective 1: Euclidean length}
\par
\begin{figure} [h!]
\includegraphics[width=0.4\textwidth]{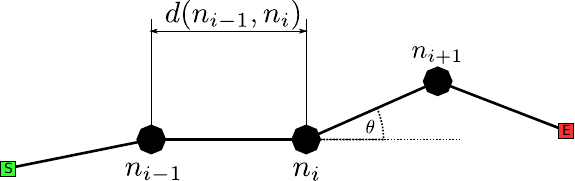}
\caption{Objectives (1) and (5) on an example path, which is modeled using a graph with nodes and edges.} \label{fig:objectives}
\end{figure}

The Euclidean length represents the distance between the start $n_S$ and the end $n_{End}$ of a path. 
It is calculated by the sum of the Euclidean distances $d(n_{i-1}, n_i)$ between the neighbouring vertex pairs $n_{i-1}$ and $n_i$ in a solution path $N$ as follows: 
\begin{equation}
\label{eq:length}
f_1(N)=\sum_{i=1}^{K-1}d(n_{i},n_{i+1})
\end{equation}
\begin{equation}
d(u,v)=\norm{u-v}_2
\end{equation}
We consider that $i=1$ corresponds to the starting point $n_S$ and the last node of a path $n_K$ maps to the endpoint denoted $n_{End}$.
Figure \ref{fig:objectives} illustrates an example. In real-world applications, this objective can be additionally used to estimate fuel consumption. 

\par
\noindent
\textbf{Objective 2: Delays}
\label{par:acc}
\par
The second objective is meant to measure the amount of delay in a given path. In real-world applications, delays are caused by accidents or traffic. Therefore, a delay is the likelihood of having an accident on each node of the path. 
The delay per path segment between the nodes $n_i$ and $n_{i+1}$ is defined by the differences between the corresponding values of the two adjacent nodes. Our proposed second objective $f_2$ calculates the sum of $delay$ for all the edges on a given path $N$:
\begin{subequations}
 \begin{equation}
\label{eq:accidents}
f_2(N)=\sum_{i=1}^{K-1}delay(n_i,n_{i+1})
\end{equation}
\begin{equation}
\begin{split}
&delay(n_i,n_{i+1})= \\
&\left\lbrace \begin{array}{cl}
2 & \;\textrm{if}\;\;v\left(n\right)\not= v\left(n_1 \right)\\
3 & \;\textrm{if}\;\;v\left(n\right)=\mathrm{city}\wedge v\left(n_1 \right)=\mathrm{city}\\
1 & \;\textrm{if}\;\;v\left(n\right)=\mathrm{country}\wedge v\left(n_1 \right)=\mathrm{country}\\
\frac{1}{5} & \mathrm{\ otherwise}
\end{array}\right.
    \end{split}
\end{equation}
\end{subequations}

\par
\noindent
\textbf{Objective 3: Elevation}
\par
\begin{figure} [t]
    \centering
    \includegraphics[width=0.4\textwidth]{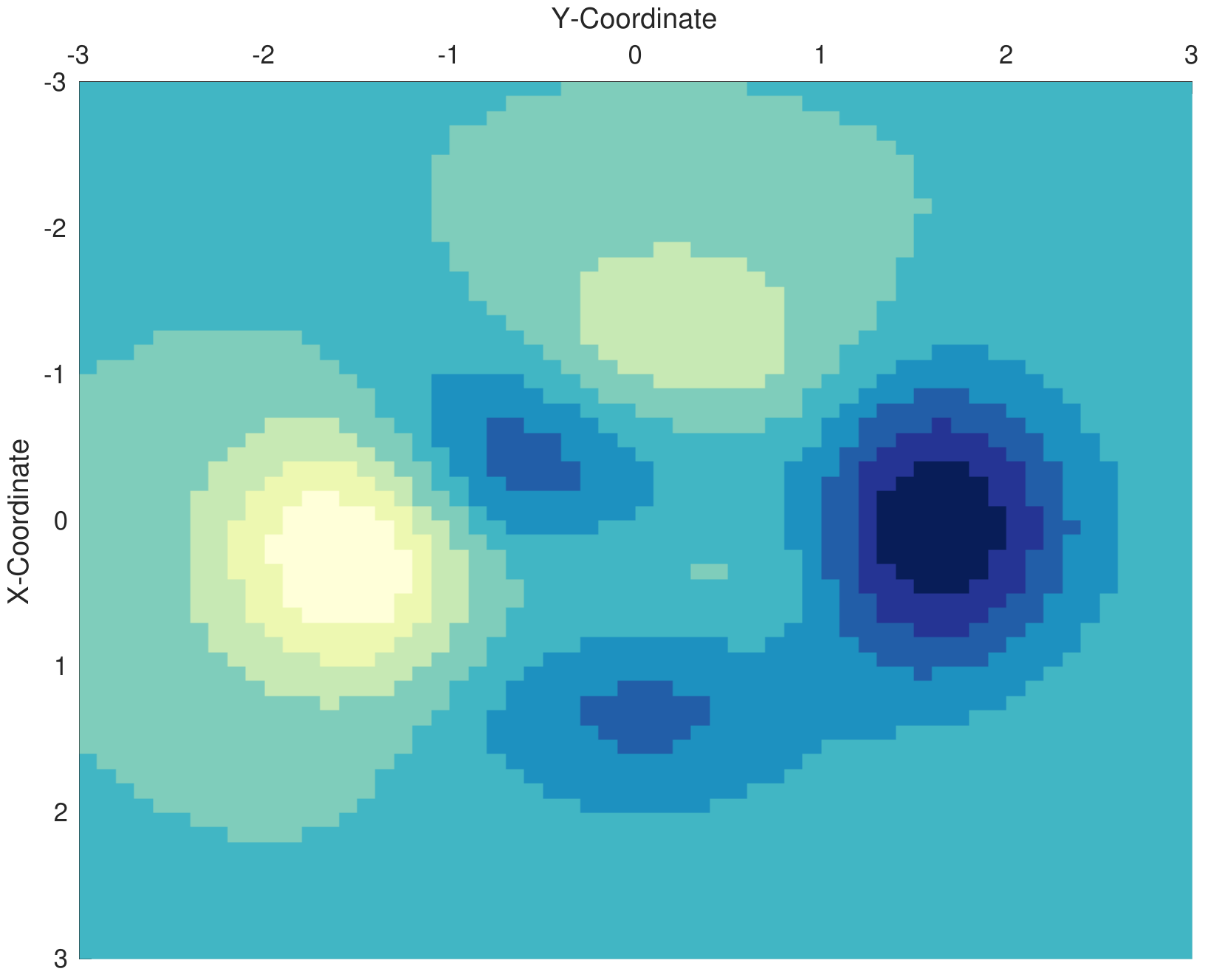}
    \caption{Example of elevation function $h$ illustrated by contours. Dark colors illustrate the amount of elevation}
    \label{fig:objective3}
\end{figure}

The aggregated ascent of a solution path is represented in the third objective. Similar to the second objective, this objective is also defined on a grid map. Our proposed benchmark contains various possibilities for defining the elevation function $h(n_i)$ which is defined on a node $n_i$. Examples of the elevation function are shown in \cref{fig:objective3}.
The ascent is calculated between two nodes in the graph $e(n_i,n_{i+1})$. 
and the third objective $f_3(N)$ is the sum of the elevations between all the nodes in a path $N$: 
\begin{equation}
\label{eq:elevation}
\begin{split}
f_3(N)& =\sum_{i=1}^{K-1}e(n_i,n_{i+1}) \\
e(m,n) & = \begin{cases} h(m)-h(n), & \mbox{if } h(m)>h(n)\\
0, & \mbox{else }
\end{cases} 
\end{split}
\end{equation}

Similar to objective 1, this objective also represents the amount of fuel consumption.
Figure \ref{fig:objective3} shows an example of the elevation function.

\par
\noindent
\textbf{Objective 4: Traveling time}

The fourth objective represents the traveling time. For this purpose, we utilize the velocity defined by $v_{max}(n_i)$ for each node $n_i$ and use the length of the path which was also used for the Objective 1: 

\begin{equation}
\label{eq:time}
f_4(N)=\sum_{i=1}^{K-1}\frac{2d(n_{i},n_{i+1})}{v_{max}(n)+v_{max}(n_{i+1})}
\end{equation}

\par
\noindent
\textbf{Objective 5: Smoothness}

The smoothness, or curvature, of a path is modeled in the fifth objective. We measure smoothness by calculating the angle between three nodes on a path, as shown in Figure \ref{fig:objectives}. The angle $\theta$ is obtained by extending the line between two nodes and measuring the angle to the third node. Similar to ~\cite{Oleiwi2014ModifiedPlanning,Jun2010Multi-objectiveAlgorithm}, we invert $a \cdot b = \norm{a}\norm{b}cos(\theta)$: 
\begin{equation}
\label{eq:smoothness}
f_5(N)=\sum_{i=2}^{K-1}arccos(\frac{\overrightarrow{n_{i}n_{i-1}}\cdot\overrightarrow{n_{i+1}n_{i}}}{\abs{\overrightarrow{n_{i}n_{i-1}}}\cdot\abs{\overrightarrow{n_{i+1}n_{i}}}})
\end{equation}
Since we intend to minimise the objective values, the smaller smoothness value represents a more straight path.

\section{Benchmark Test Suite}
Given the above model, we can generate several problem instances. In the following, we propose various examples for a test suite by selecting specific features for the defined variables of the benchmark. 
We set 4 various kinds of cells with velocity values $v_{max}(Highway) = 130\frac{km}{h}$, $v_{max}(Country) = 100\frac{km}{h} $ and $v_{max}(City) = 50\frac{km}{h}$.
As for obstacle cells ($v_{max} = 0$), we propose two different forms: 1) the chequerboard pattern is designed to simulate block-like environments, and 2) the lake obstacle denotes a larges region which is not passable. 
For the chequerboard obstacles, every second cell is defined as an obstacle in both $x$ and $y$ directions. The lake obstacle is defined as a circle on the grid. The circle radius is defined by a fraction of the $x$-size of the grid. \Cref{eq:checkerboardconstraint,eq:lakeconstraint} describe the obstacle as a variant of the square wave function and circle function, respectively. In the tested instances, the Lake obstacles are defined by a radius of $0.25 x_{max}$.
\Crefrange{fig:map:lake}{fig:map:checkboard} shows the two obstacle types on an example instance of the benchmark problem.
\Cref{fig:map:streets} shows an example instance of size 20. The darkest grid cells represent large speed grids (motorways), the lighter colored cells represent paths with smaller velocity values and the lightest colors indicate the obstacles with $v_{max} = 0$.

\begin{equation}
\label{eq:streets}
\begin{split}
v_{max}(x,y) & = \begin{cases} v_H, & \mbox{if } w(x,y)>0.9\\
v_S, & \mbox{if } w(x,y)<-0.4\\
v_M, & \mbox{else} 
\end{cases} \\
w(x,y) & = \mathrm{max}\left(sin(x),cos(y)\right)
\end{split}
\end{equation}

\begin{subequations}
\begin{equation}
\label{eq:checkerboardconstraint}
\begin{split}
    g_{CH}(x,y) =& \mathrm{sign}\left(\mathrm{sin}\left(\frac{\pi }{2}+\pi \,x\right)\right)+\mathrm{sign}\left(\mathrm{sin}\left(\frac{\pi }{2}+\pi \,y\right)\right)\\
    &-2\,\mathrm{\Pi}\left(x-x_{max}\right)\,\mathrm{\Pi}\left(y-y_{max}\right) < 2
\end{split}
\end{equation}
\begin{equation}
\begin{split}
    &\Pi(x) = H\left(x+\frac{1}{2}\right)-H\left(x-\frac{1}{2}\right)\\
    & \text{where } H(x) \text{ is the Heaviside step function}
\end{split}
\end{equation}
\end{subequations}

\begin{subequations}
\begin{equation}
\label{eq:lakeconstraint}
\begin{split}
    g_{LA}(x,y) &= \\
    &{{\left(x-\frac{x_{max}}{2}\right)}}^2 +{{\left(y-\frac{y_{max}}{2}\right)}}^2 -{\left(r \,x_{max} \right)}^2 < 0\\
    & \text{where } r \text{ denotes the radius ratio}
\end{split}
\end{equation}
\end{subequations}

As for the elevation, we set four hill functions in the domain $(-3,3)$ which will be scaled when applied to the grid with cell coordinates $(x, y)$ represented by the node $n$ in the path segment:
\begin{equation}
\label{eq:heights}
\begin{split}
h_{1}(x,y)=& 3 (1 - x)^2 e^{-x^2 - (y + 1)^2} - 10 e^{-x^2 - y^2}\\
 &(-x^3 + x/5 - y^5) - 1/3 e^{-(x + 1)^2 - y^2} \\
h_2(x,y)=& 5e^{-(x + 1.5)^2 - (y + 1.5)^2}\\
h_3(x,y)=& 5e^{-(x - 1.5)^2 - (y - 1.5)^2}\\
h_4(x,y)=& 5e^{-(x - 1.5)^2 - (y + 1.5)^2}
\end{split}
\end{equation}
These functions can be linearly combined which can produce many variations for the elevation function $h$: 
\begin{equation}
h(x, y) = \sum_{i=1}^{nh} h_i, ~~~~nh = \{1, 2, 3, 4\} \\
\end{equation}
For the third objective, we aggregate positive slopes, as we want to focus on flat routes. Taking also negative elevations into account could result in a path containing a hill with a high steep which would not be beneficial to bulky goods transports. 

All these variations of the properties are used in the name of a benchmark instance. The name starts with \emph{ASLETISMAC} for the five objectives to be minimised (Ascent, Length, Time, Smoothness and Accidents), then the obstacle type, followed by the size in X and Y directions, then the elevation function is represented (PM stands for the peaks-function $h_1$ and the combination is set to P$nh$), followed by the $2^k$-neighbourhood and the backtracking property (B followed by T for True or F for False). For example, \emph{ASLETISMAC$\_$CH$\_$X10$\_$Y10$\_$P1$\_$K2$\_$BF} defines an instance with the chequerboard obstacles, sized $10x10$, $nh= 1$ as the elevation function, four possible neighbours (K2\textrightarrow $2^2=4$), but no backtracking (BF).

For the values of delays in the second objective, we can refer to real-world statistical data (see \cref{eq:accidents})\footnote{\url{https://www.destatis.de/EN/Themes/Society-Environment/Traffic-Accidents/_node.html}}.

\subsection{Obtaining the true Pareto-Front}
In order to make this benchmark-framework available, we performed an exhaustive search on 289 benchmark instances with different obstacle types, sizes, elevation-functions, neighbourhood metrics but only without backtracking, since these instances were too complex to provide any meaningful algorithmic results. However, the benchmark contains these as well, and we want to investigate these instances in the future as well. We provide the corresponding 273 Pareto-Fronts and their corresponding sets for the five-objective optimisation. The missing 16 instances do not have any solution due to the nature of the benchmark, i.e. the lake obstacle covers too much of the area. Depending on the type, there are different numbers of Pareto-optimal solutions. In order to obtain the fronts, we performed a depth-first search (DFS) from the cell at the northern-west corner to the south-east corner cell. The larger the instances are, the longer the DFS takes to complete. The most complex in terms of the number of possible paths, which we evaluated, is the instance 
\emph{ASLETISMAC$\_$NO$\_$X14$\_$Y14$\_$PX$\_$K3$\_$BF}, that has a size of 14x14, 4-neighbourhood and no backtracking. For this instance, there are $1,409,933,619$ possible paths. 


We implemented an exhaustive search by running the DFS as mentioned earlier and evaluated the solutions according to the objectives specified in \cref{sec:objectives} for all the 289 instances. After the evaluations, we calculated the non-dominated set representing the true Pareto-front (provided together with the codes in supplementary materials).

\section{Experiments}
\label{sec:experiments}
In the experiments, we aim to investigate the many-objective pathfinding problem for one instance of the benchmark and provide an in-depth analysis of the proposed objective functions. We perform three different state-of-the-art evolutionary algorithms on several instances to evaluate the complexity of the benchmark. Furthermore, we present a custom mutation operator, which can operate on a variable-length chromosome consisting of a list of nodes.

\subsection{Search space encoding and operators}
In our proposed benchmark, we consider a solution to be a sequence of nodes $N=(n_1,\cdots,n_K)$ with a variable-length $K$. We take this representation for the encoding in evolutionary algorithms. The variable-length chromosome poses difficulties for the algorithms but can be very efficient when using realistic data since intersections and endpoints are not homogeneously distributed, and paths usually have different lengths. This representation has been used by \cite{Lamini2018GeneticPlanning,Tuncer2012DynamicAlgorithm,Li2006AnRobots} and studied by~\cite{Beke2020AMultigraphs}.


We use a one- or two-point cross-over for this encoding as follows. If two selected solutions have intersection points except for the start and end nodes, these points can be used as possible cut-off points. If there are fewer than two intersections, we use a one-point cross-over.
Additionally, we define the so-called \emph{perimeter mutation operator}. From a given path which is to be mutated, we took two arbitrary points within a maximum network distance $d_{max}=\frac{\abs{N}}{2}$ and compute their middle point. We then search for a random point in the network within a maximum distance of $r_{max}$, using an R-Tree index from that point. Then we perform a random-search (local search) from the first and second points to it. 
Depending on the benchmark instance, we either consider all neighboring nodes within the radius in positive cardinal and diagonal directions (instances of type \emph{K3,BF}) or a subset of them: nodes in positive cardinal directions for \emph{K2,BF}. 




\subsection{Experiment setup}
In the experiments, we use the NSGA-II \cite{Deb2002ANSGA-II}, NSGA-III \cite{Deb2014AnConstraints} and DIR-enhanced NSGA-II (d-NSGA-II)~\cite{Cai2018AOptimization} algorithms.  
The d-NSGA-II uses a diversity indicator based on reference vectors~\cite{Cai2018AOptimization}, making it suitable for many-objective optimisation problems.
For all three algorithms, we set the population size to 212 as in the original NSGA-III study. We set the probabilities for cross-over and mutation to 0.8 and 0.2, the number of divisions for NSGA-III to $p=6$, maximum number of generations to 500, and all for 31 runs for statistical analysis. 
The task of the pathfinding algorithm is to find a path from the north-west corner to the south-east corner. 
We take 273 problem instances:


We implemented the experiments using the JGraphT library~\cite{jgrapht2019} for graph storage and the jMetal framework, version 6.0 (development snapshot), for the algorithm's execution~\cite{Nebro2015RedesigningFramework}.
To compare the algorithms, we calculated the \igdp indicator~\cite{Ishibuchi2015AIndicator,Ishibuchi2014DifficultiesProblems}. The results are compared and tested for statistical significance using the two-sided pairwise Mann-Whitney-U Test, with the null hypothesis that the distributions of the three samples have equal medians. Statistical significance of the differences between the performance is assumed for a p-value smaller than $0.01$.\\

\subsection{Results}
\begin{figure}[t]
\centering
\includegraphics[width=0.9\linewidth]{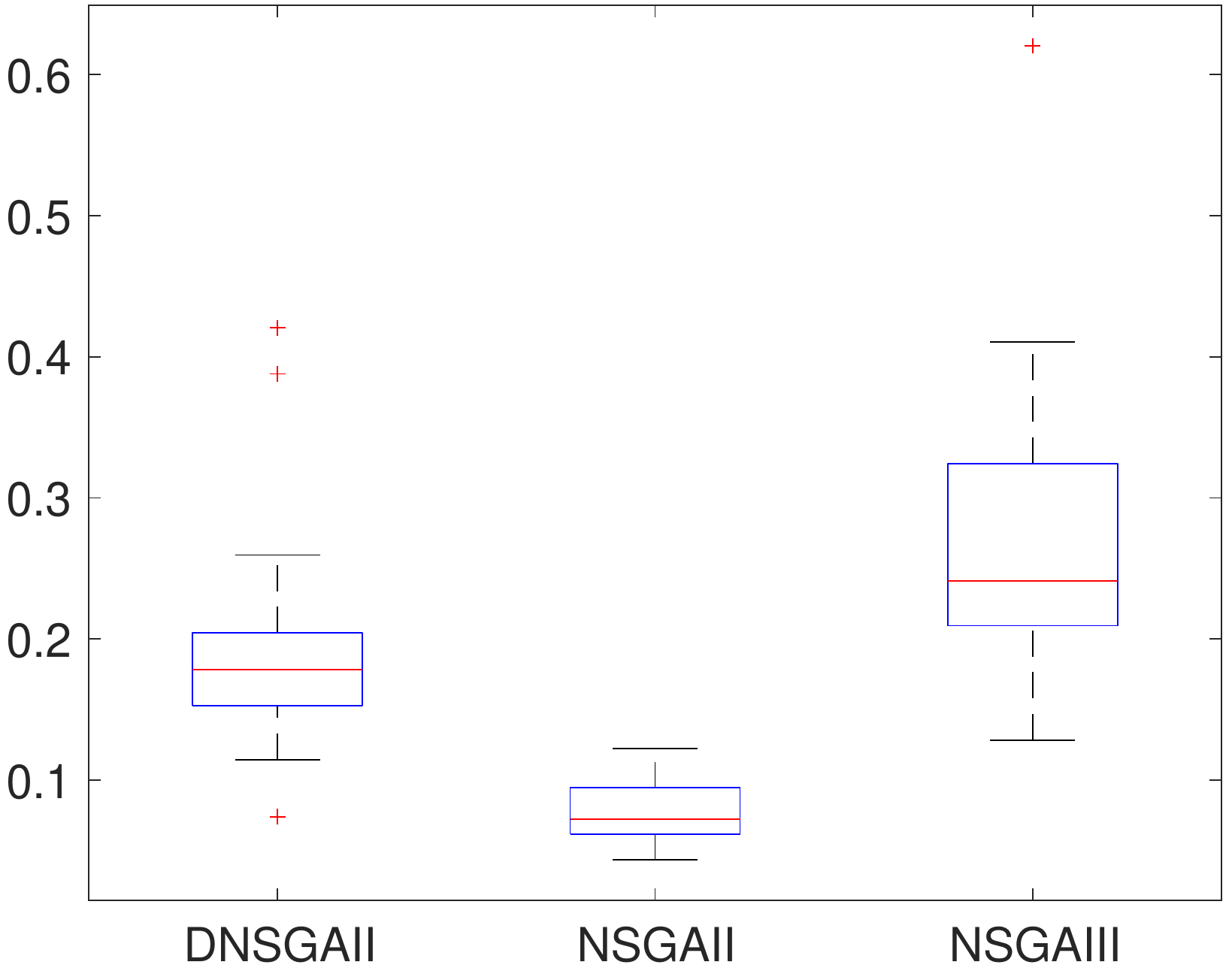}
\caption{Obtained IGD$^+$ Values for the instance ASLETISMAC$\_$NO$\_$X14$\_$Y14$\_$PM$\_$K3$\_$BF} 
\label{fig:result:igd}
\end{figure}

In the first part of our analysis, we count the number of successful runs in which the algorithms could obtain the entire Pareto-front. A front is found if the \igdp is $0$ in all 31 runs on the algorithms. Given 273 valid instances, NSGA-II, NSGA-III and d-NSGA-II were not able to find the front for 235, 234 and 241 instances. This indicates the difficulty of the benchmark for certain instances. 
\Cref{fig:result:igd} shows the obtained \igdp values for the instance ASLETISMAC$\_$NO$\_$X14$\_$Y14$\_$PM$\_$K3$\_$BF for which none of the algorithms found the whole Pareto-front, indicating complexity of the problem. We observe that NSGA-II obtains the best result, even if NSGA-II is not the best option for many-objective problems. 

\begin{table}
\centering
\resizebox{\linewidth}{!}{%
\begin{tabular}{cccc}
\toprule
 & DNSGAII & NSGAII & NSGAIII \\
\midrule
NO$\_$X14$\_$Y14$\_$PM$\_$K3$\_$BF & 0.17825 (0.051749)* & \textbf{0.072188} (0.032839) & 0.24099 (0.11487)* \\
\bottomrule
\end{tabular}}
\caption{Obtained median and IQR values for the \igdp indicator on the different algorithms. Best performance is shown in bold. An asterisk (*) indicates statistical significance compared to the respective best algorithm.}
\label{table:tab:results:example}
\end{table}

\begin{figure*}[t]
     \centering
\subfloat[\igdp Values for all P1 instances]
{
\label{fig:results:p1}
\includegraphics[width=0.45\textwidth]{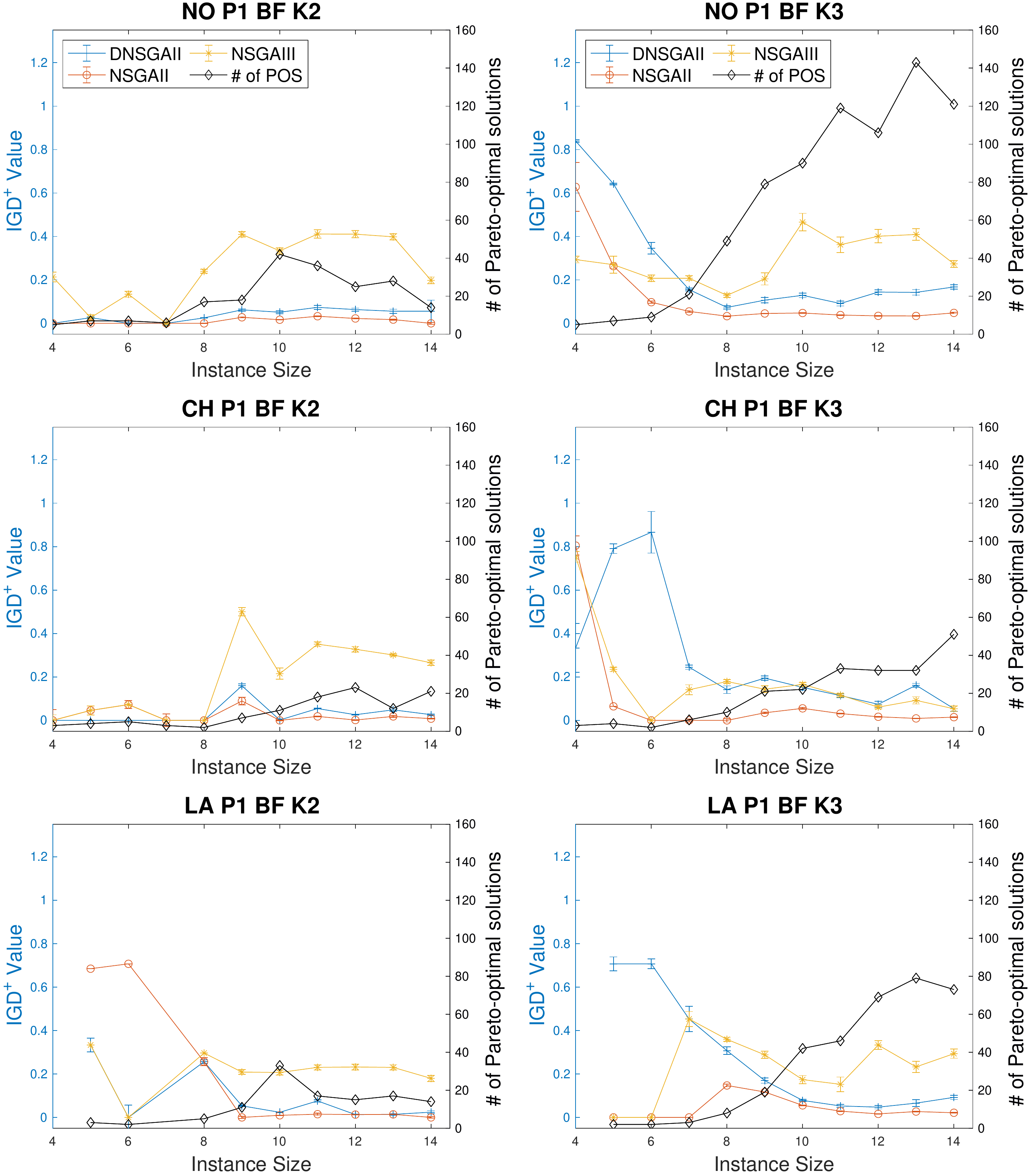}
}\quad
\subfloat[\igdp Values for all P2 instances]
{
\label{fig:results:p2}
\includegraphics[width=0.45\textwidth]{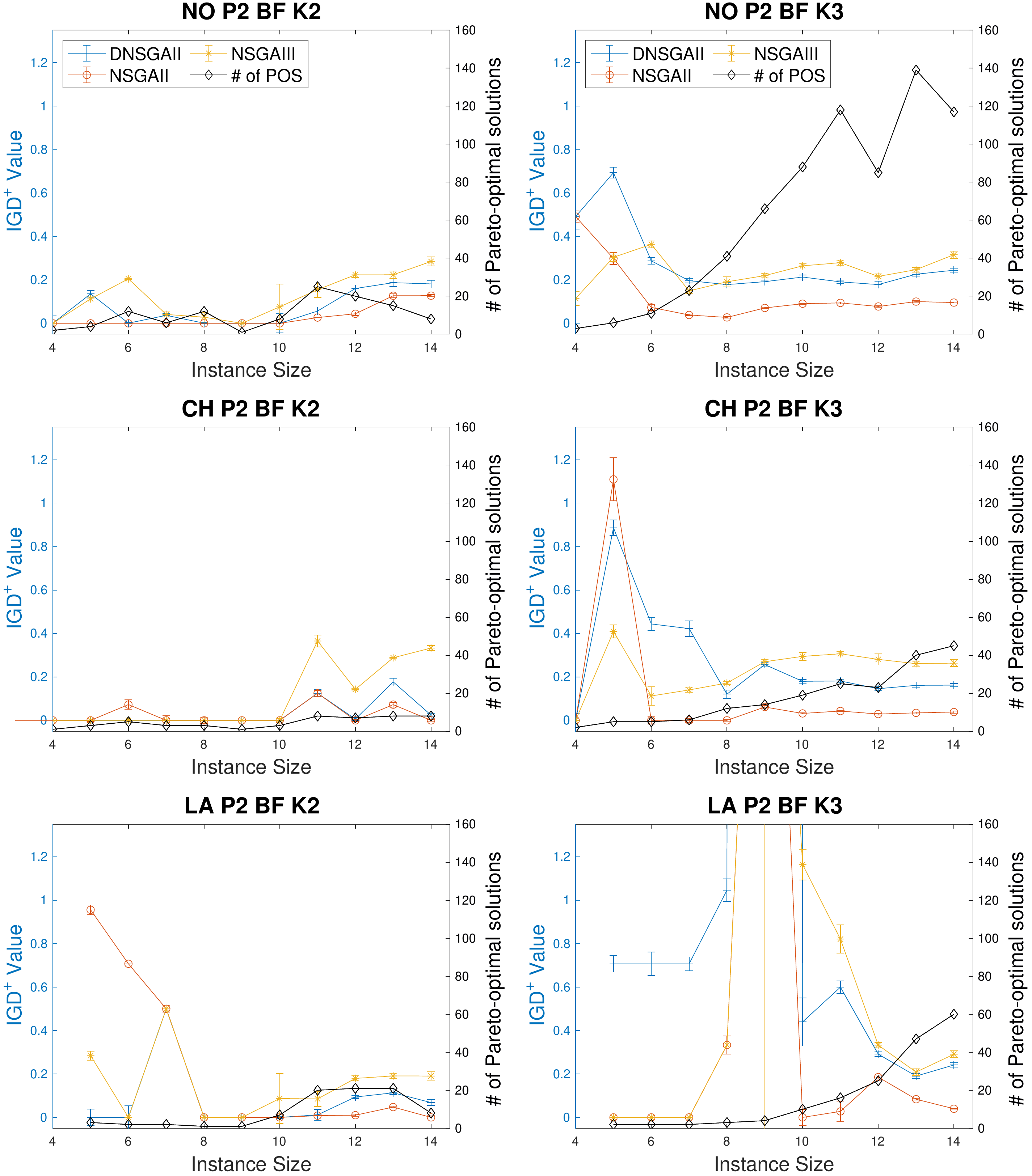}
}\\
\subfloat[\igdp Values for all P3 instances]
{
\label{fig:results:p3}
\includegraphics[width=0.45\textwidth]{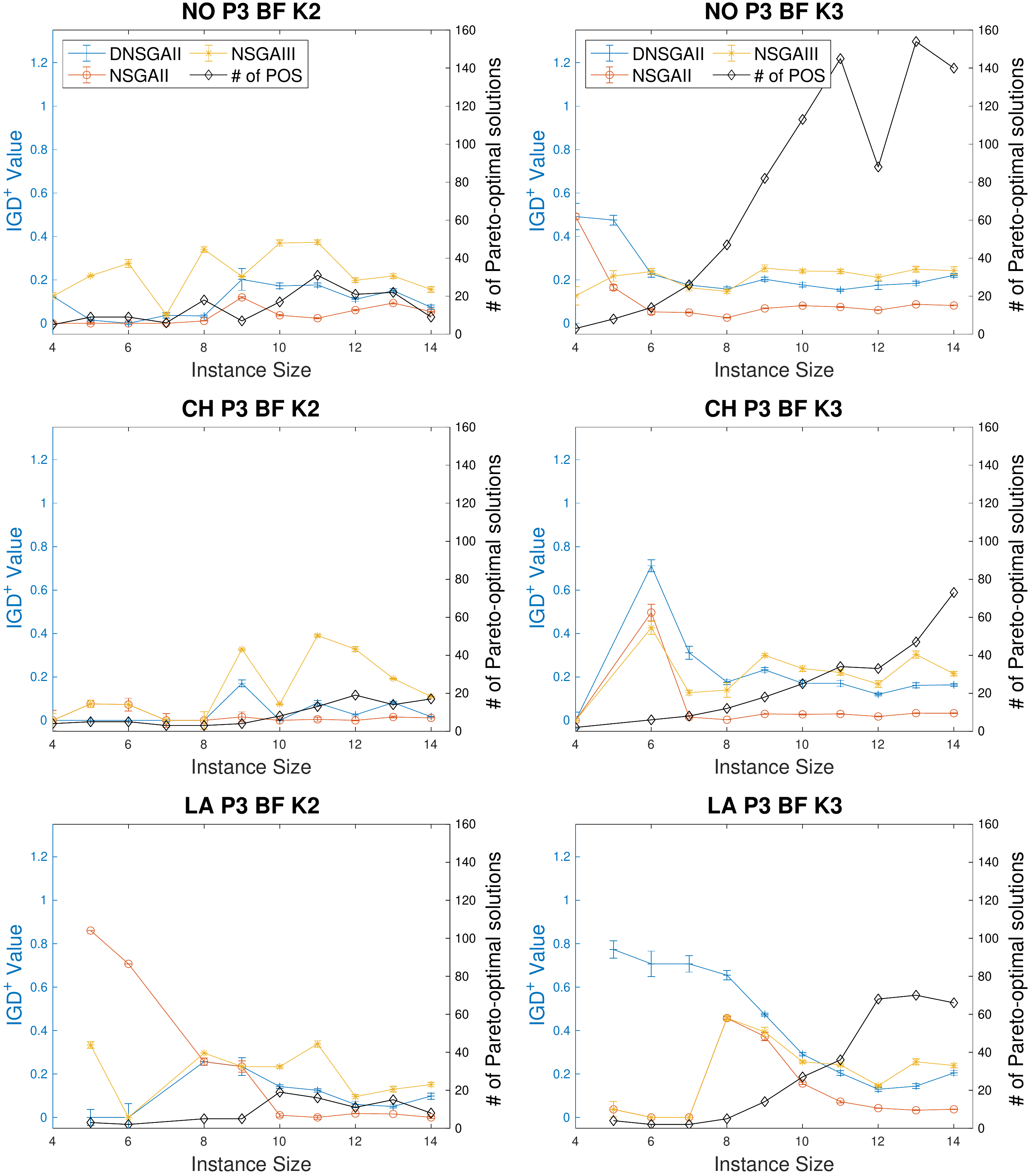}
}\quad
\subfloat[\igdp Values for all PM instances]
{
\label{fig:results:pm}
\includegraphics[width=0.45\textwidth]{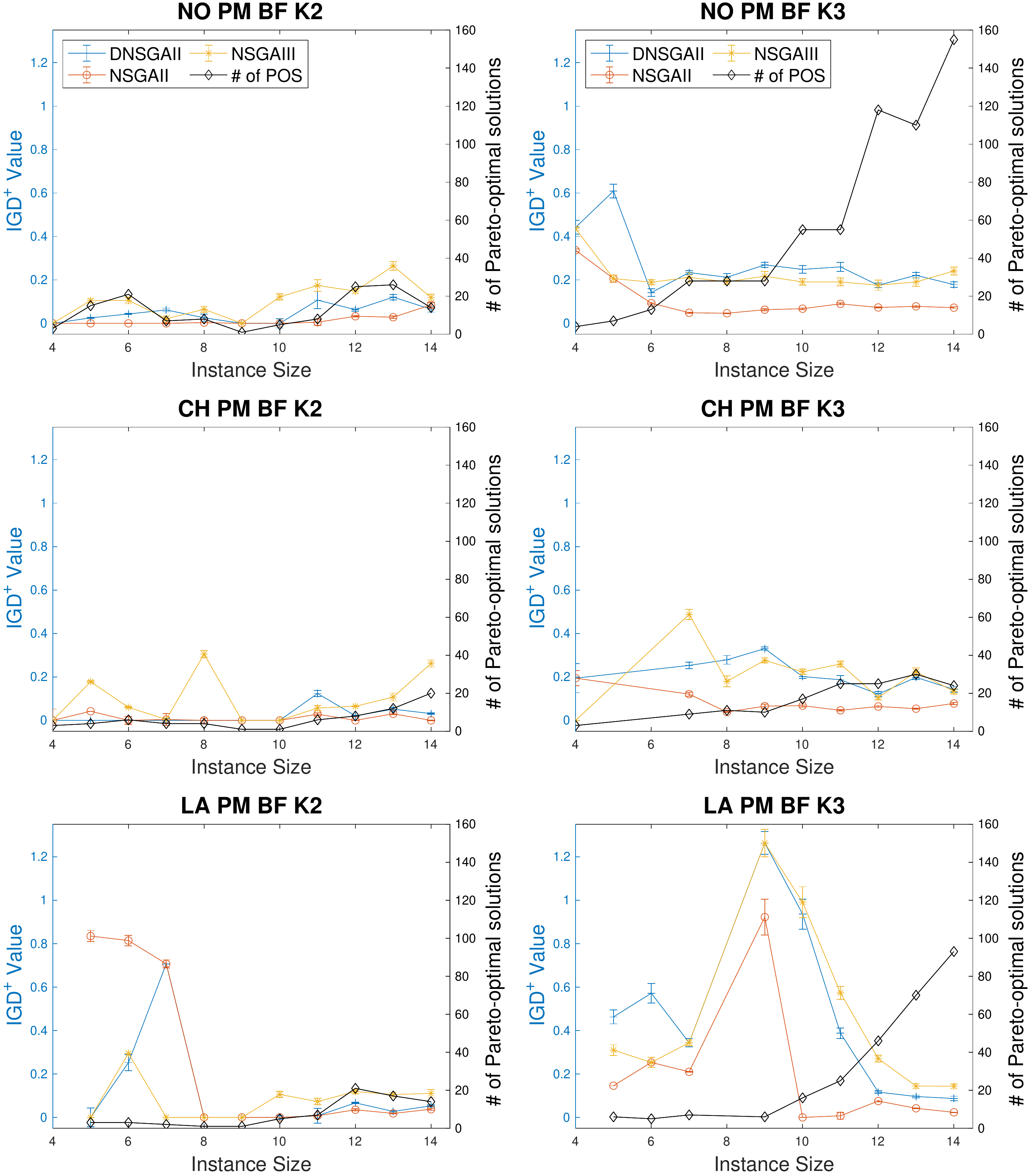}
}%
\caption{The obtained \igdp values with respect to the different type, ordered by instance size.}\label{fig:results:igdHeight}

\end{figure*}

Overall, NSGA-II performed the best in the \igdp indicator on the majority of instances (statistically significant difference for $p<0.01$, see \cref{fig:results:igdHeight}), which can be due to the crowding distance estimation to maintain diversity which is beneficial to irregular Pareto-fronts~\cite{Cai2018AOptimization}.\\ We provide all obtained graphs in the supplementary material for further analysis.


\subsection{Real-World Data}
\label{sec:realworld}
\begin{figure}[t]
\includegraphics[width=0.45\textwidth]{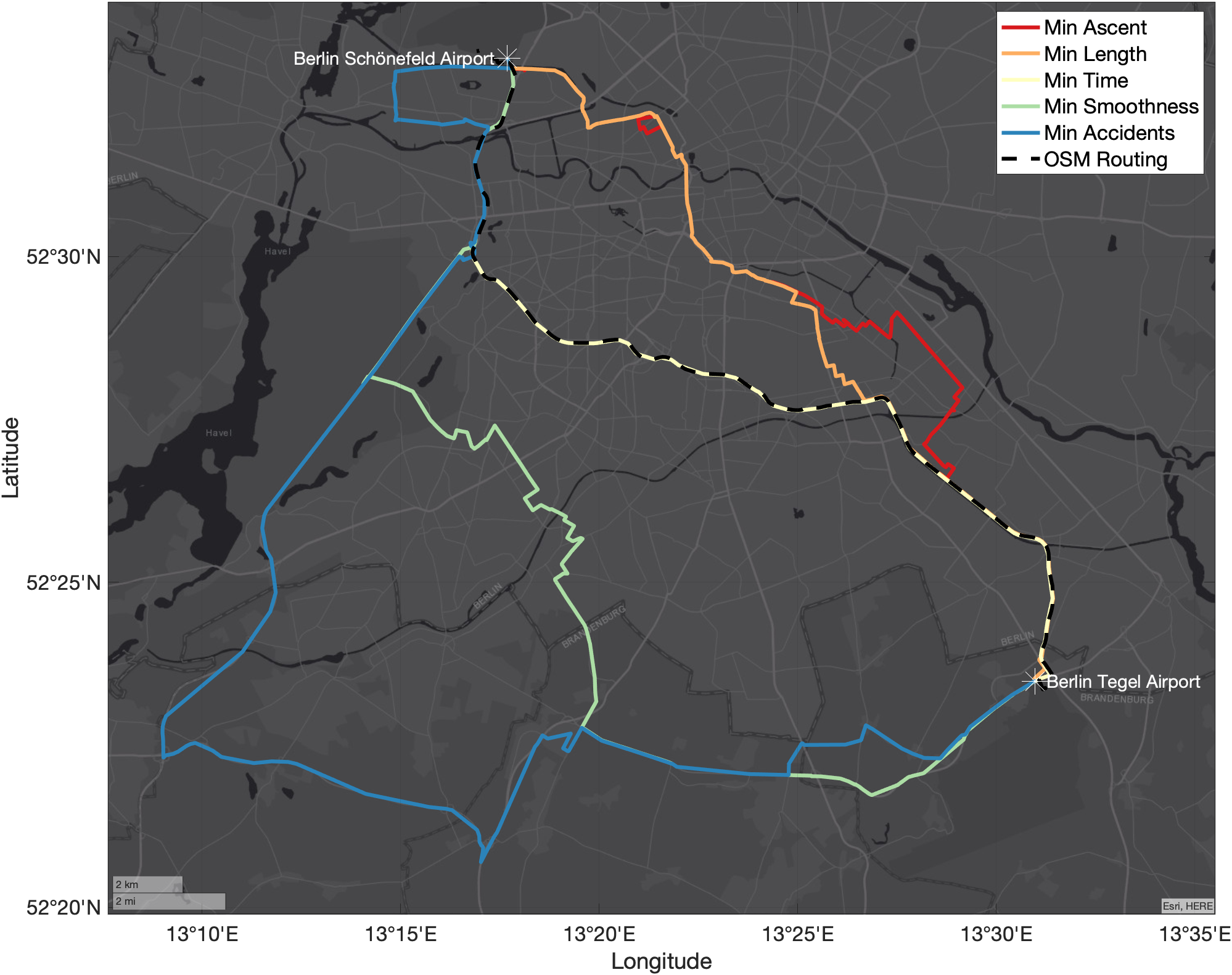}
\caption{Map of Berlin showing the best path in terms of each objective. \plotbar{1} Min Ascent, \plotbar{2} Min Length, \plotbar{3} Min Time, \plotbar{4} Min Smoothness, \plotbar{5} Min delay, the dashed black line represents the route from the original OpenStreetMap Routing Service}
\label{fig:exp:berlin}
\end{figure}

In the following, we aim to transfer the problem from the proposed benchmark to a real-world application. We use the data on the map of Berlin and compute a set of paths between the two airports \textit{Berlin-Tegel} and \textit{Berlin-Sch\"onefeld}. For this purpose, we use OpenStreetMap data which we imported and converted to an undirected graph via the \textit{osmnx} library \cite{Boeing2017OSMnx:Networks}. We simplify the network by removing nodes which do not represent an intersection. The resulting graph has $63731$ vertices and $84912$ edges. For merged edges, we took the maximum values of the merged partners and aggregated the distances. Due to this, our computed path is an approximation but can be used to analyse the algorithm's performance on real-world data. 
\Cref{fig:exp:berlin} shows the layout of the map and depicts the start and endpoint. 

\begin{table}
\centering
\resizebox{\linewidth}{!}{%
\begin{tabular}{cccc}
\toprule
 & DNSGAII & NSGAII & NSGAIII \\
\midrule
REALMAP & 0.17849 (0.082822)* & 0.11123 (0.056836) & \textbf{0.11023} (0.049451) \\
\bottomrule
\end{tabular}}
\caption{Obtained median and IQR values for the IGD+ indicator on the different algorithms. Best performance is shown in bold. An asterisk (*) indicates statistical significance compared to the respective best algorithm.}
\label{table:tab:IGD+}
\end{table}

The OpenStreetMap provides the GPS-coordinates for a grid representation which can be easily used to measure the path length for the first objective. As for the second objective concerning the delay (number of accidents), we used the publicly available accident statistic data\footnote{ \url{https://unfallatlas.statistikportal.de/_opendata2019.html}} and mapped them to the imported network. Since the coordinates of the accidents are mostly different from the available nodes in the network, we defined an R-Tree-Index on the network and performed a nearest node search for each accident. In this way, we aligned each accident to a node in the network. 
The third objective was measured using the Google Maps Elevation API\footnote{\url{https://developers.google.com/maps/documentation/elevation/start}}. The elevation is obtained in meters over the sea level and written to the node's properties.
For the smoothness, we simplified the network to straight connections between nodes. Therefore, it is obtained in the same way as in the proposed benchmark.
From the OpenStreetMap network, we could also obtain the information about speed limits per street segment. We calculated the time needed per segment as the ratio of distance and speed. Summing up the values of each segment results in the total traveling time (Objective 5).
For the experiments, we take the same parameter settings as above with only one-point cross-over. 

\begin{figure}[t]
\centering
\includegraphics[width=0.9 \linewidth]{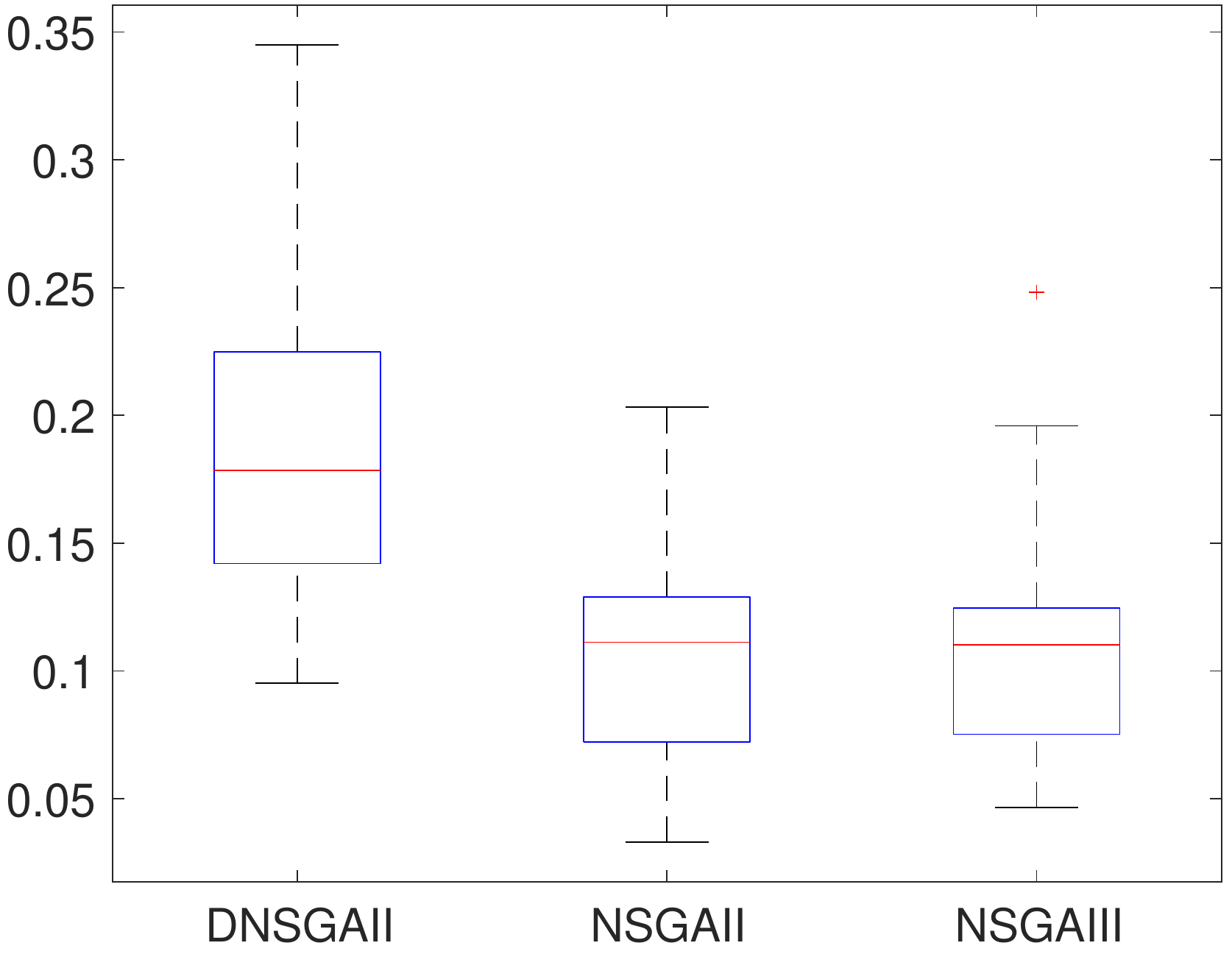}
\caption{Obtained IGD$^+$ values on Real-world problem} 
\label{fig:result:realmap:igdplus}
\end{figure}

\begin{figure}[t]
\centering
\includegraphics[width=\linewidth]{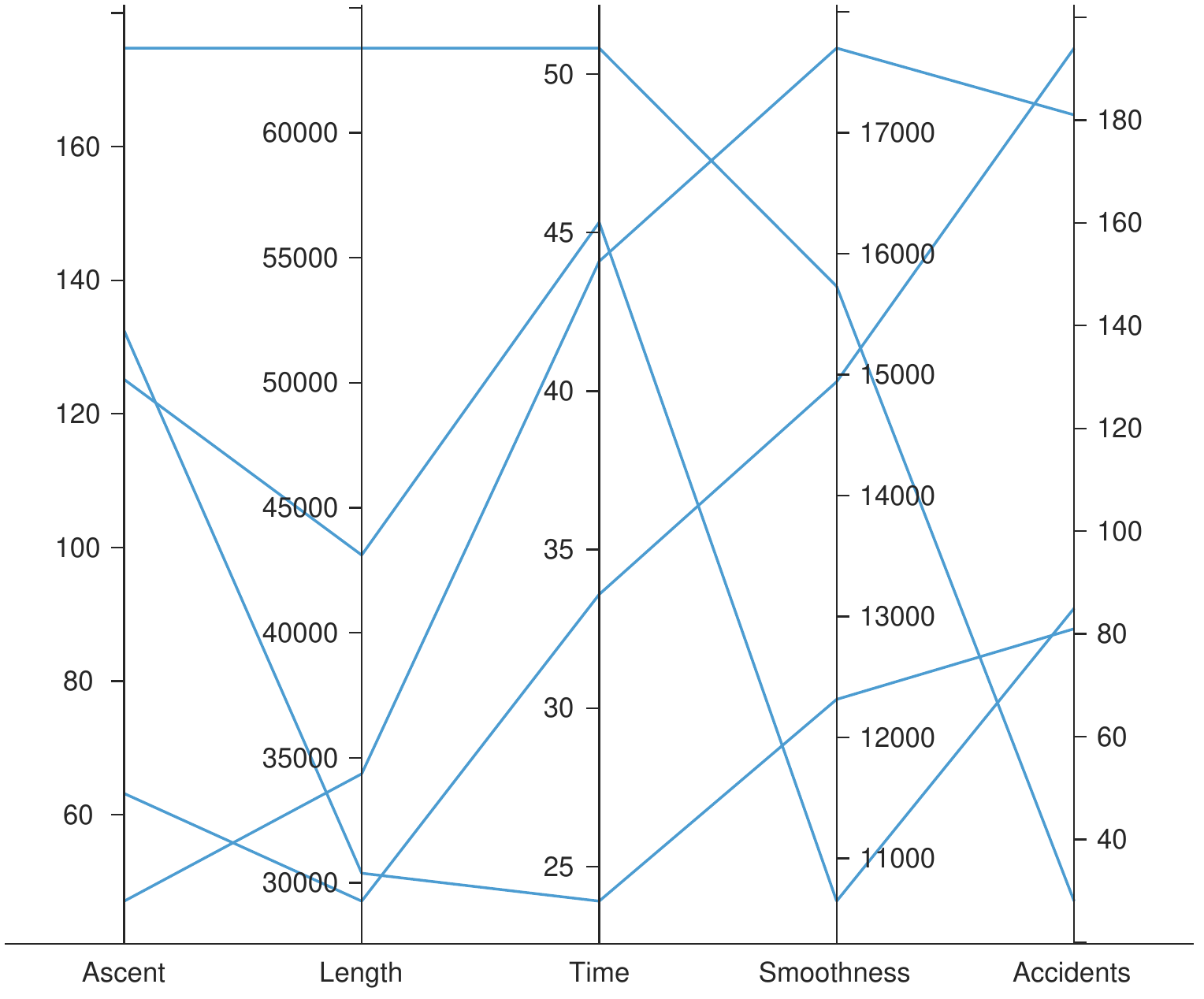}
\caption{Parallel coordinates plot of the best paths per objective.}
\label{fig:result:realmap:parallel}
\end{figure}

Since this is a real-world problem, we do not know the true Pareto-front. In order to approximate the performance of the algorithms, we combined all results from all three algorithms and all 31 runs and calculated the non-dominated solution set. We obtained 1422 non-dominated solutions.
\Cref{fig:exp:berlin} shows a subset of the obtained non-dominated solutions and the route obtained from the OpenStreetMap routing service. For clarity, we do not depict the whole set, as the number of non-dominated solutions usually increases with an increasing amount of objectives. The figure shows five non-dominated routes from one airport to the other, representing the best solution per objective. It is visible that the routes have differences. Furthermore, the route of the least accidents is mostly going over highways, indicating that the algorithms could explore the search space. Interestingly, our obtained route with the least time is the same as the obtained one from the OSM routing service.
With the obtained reference from all runs, we were able to calculate the \igdp indicator for the three algorithms. \Cref{table:realMap:wilcox:igdplus} shows the obtained results in terms of statistically significance, \cref{fig:result:realmap:igdplus} shows the respective values. \Cref{fig:result:realmap:parallel} shows the parallel plot of the best solutions per objective. 

\subsection{Benchmark vs Real-World}
\begin{figure}[ht]
     \centering
\subfloat[Real-World Data]
{
\label{fig:result:realref:ASAC}
\includegraphics[width=0.2\textwidth]{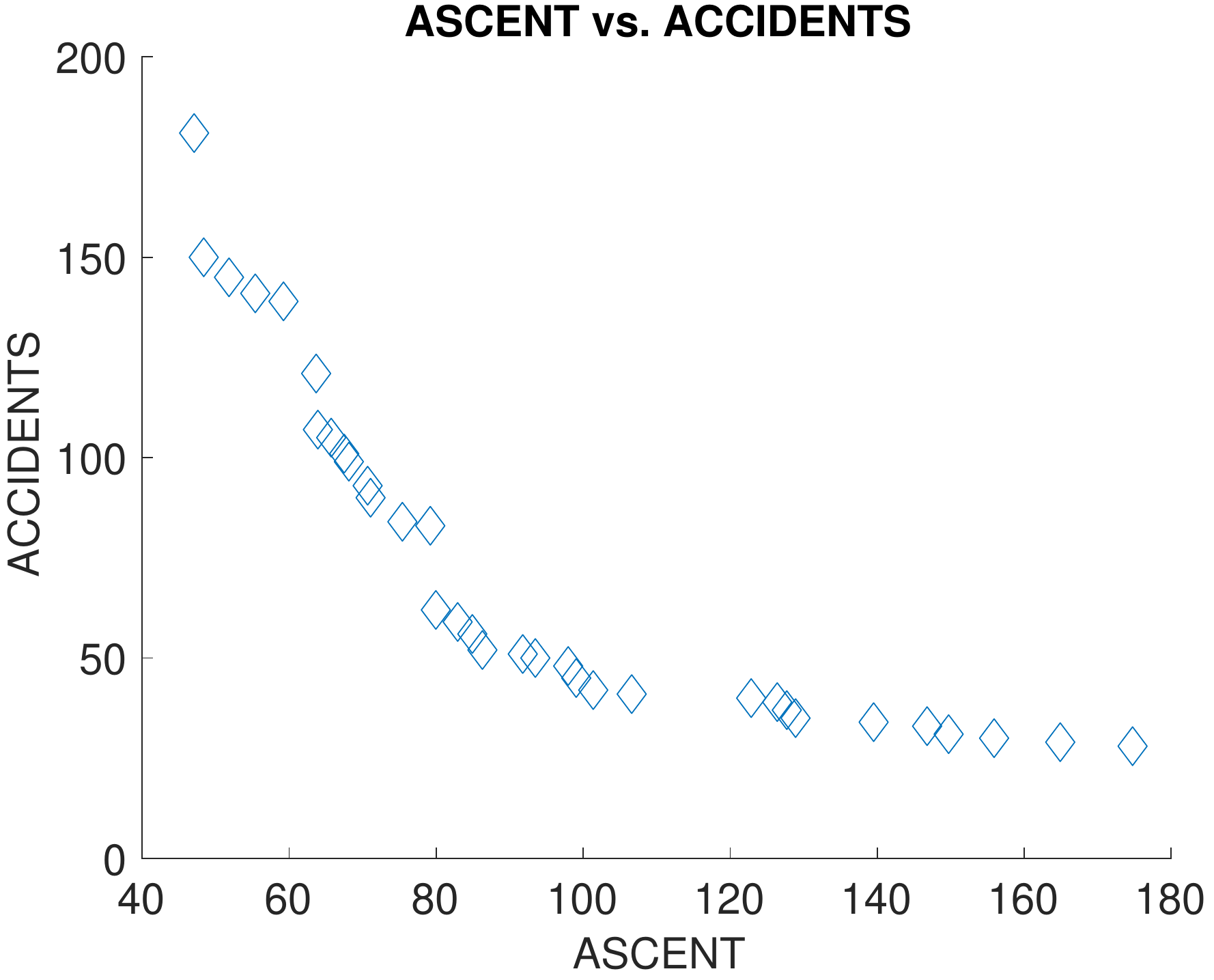}
}\qquad
\subfloat[Benchmark instance]
{
\label{fig:result:trueref:ASAC}
\includegraphics[width=0.2\textwidth]{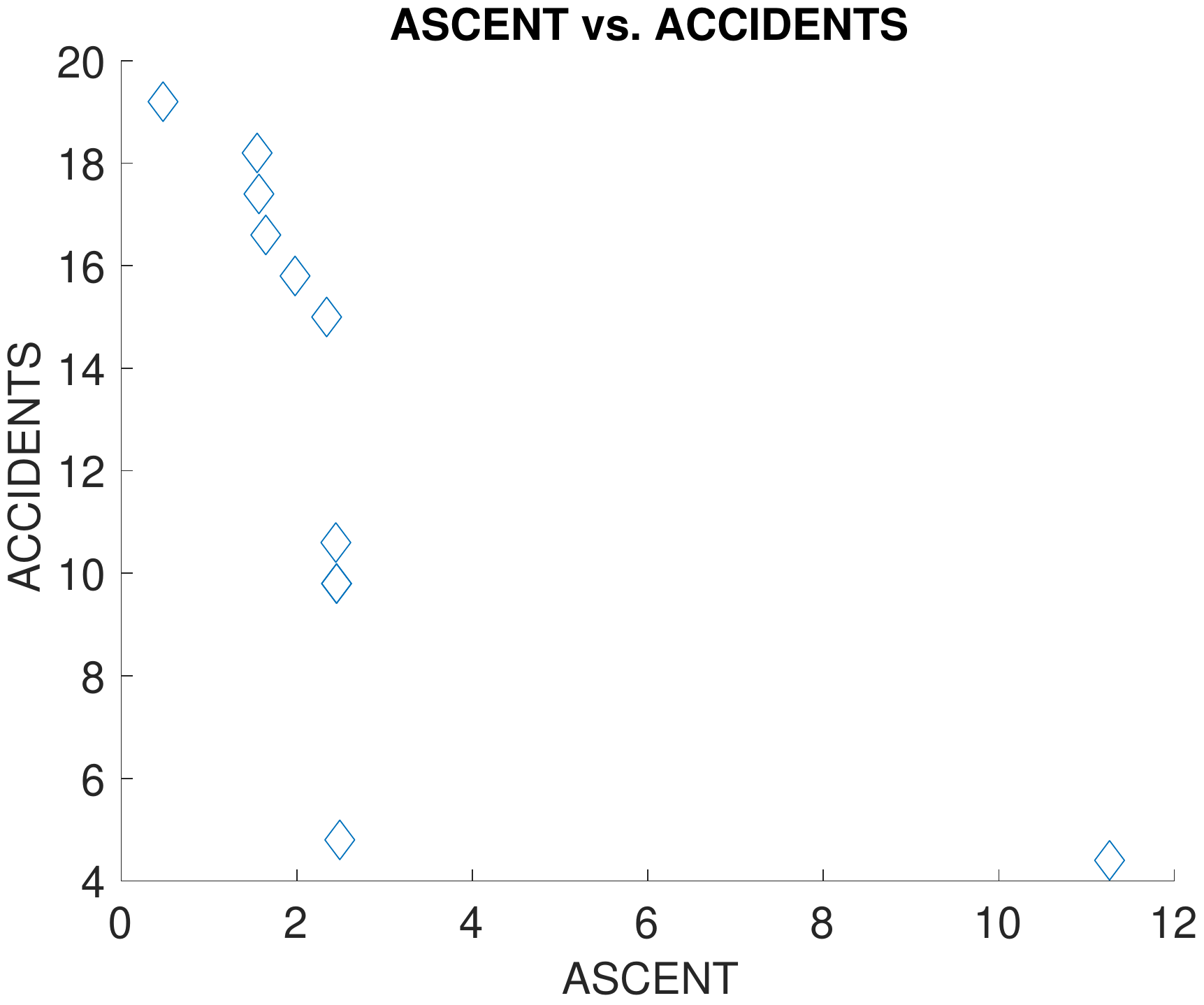}
}%
\caption{Plots of the pairwise front between the Ascent-objective and the Accidents-objective on the real-world data and the instance ASLETISMAC$\_$NO$\_$X14$\_$Y14$\_$PM$\_$K3$\_$BF}
        \label{fig:results:comparison:ASAC}
\end{figure}

We performed the three algorithms on several benchmark instances as well as on simplified real-world data. We obtained a reference front from the real-world example by combining all results and determining the non-dominated solution set. To compare our benchmark framework to real-world data and to evaluate how much it represents actual environments, we look at some plots showing the true Pareto-fronts and obtained reference fronts. \Cref{fig:results:comparison:ASAC} shows two plots of the pairwise non-dominated solution set between the objectives \emph{Ascent} and \emph{Accidents}. The shapes have similar characteristics, but the front from the benchmark instance also has at least one specific visible knee point. When comparing \emph{Length} and \emph{Accidents} is it visible, that the real-world front has clusters where the benchmark one is more distributed. The comparison of these two experiments also has several drawbacks as the benchmark instance is derived from a grid; hence from an evenly distributed environment. The real-world experiment is solely based on actual data which is mostly not evenly distributed, depending on the location which is represented. The cartesian grid-based benchmark may represent block-like environments better. To bring this more into perspective, in the analysed benchmarks, we computed a corner-to-corner path. For other start and endpoints, the fronts can look completely different.

We performed another, preferably visual, comparison with an industry used route planning software provided by the \emph{osmr}-project\footnote{\url{http://project-osrm.org/}}. The official OpenStreetMap routing service uses the project. As visible in \cref{fig:exp:berlin} the dashed line and the line of the best path in terms of time overlap entirely. This indicates that our algorithm found the best route in terms of time, assuming that the service calculates an optimal single-objective route.

\begin{figure}[ht]
     \centering
\subfloat[Length vs. Accidents Real-World Data]
{
\label{fig:result:realref:LEAC}
\includegraphics[width=0.2\textwidth]{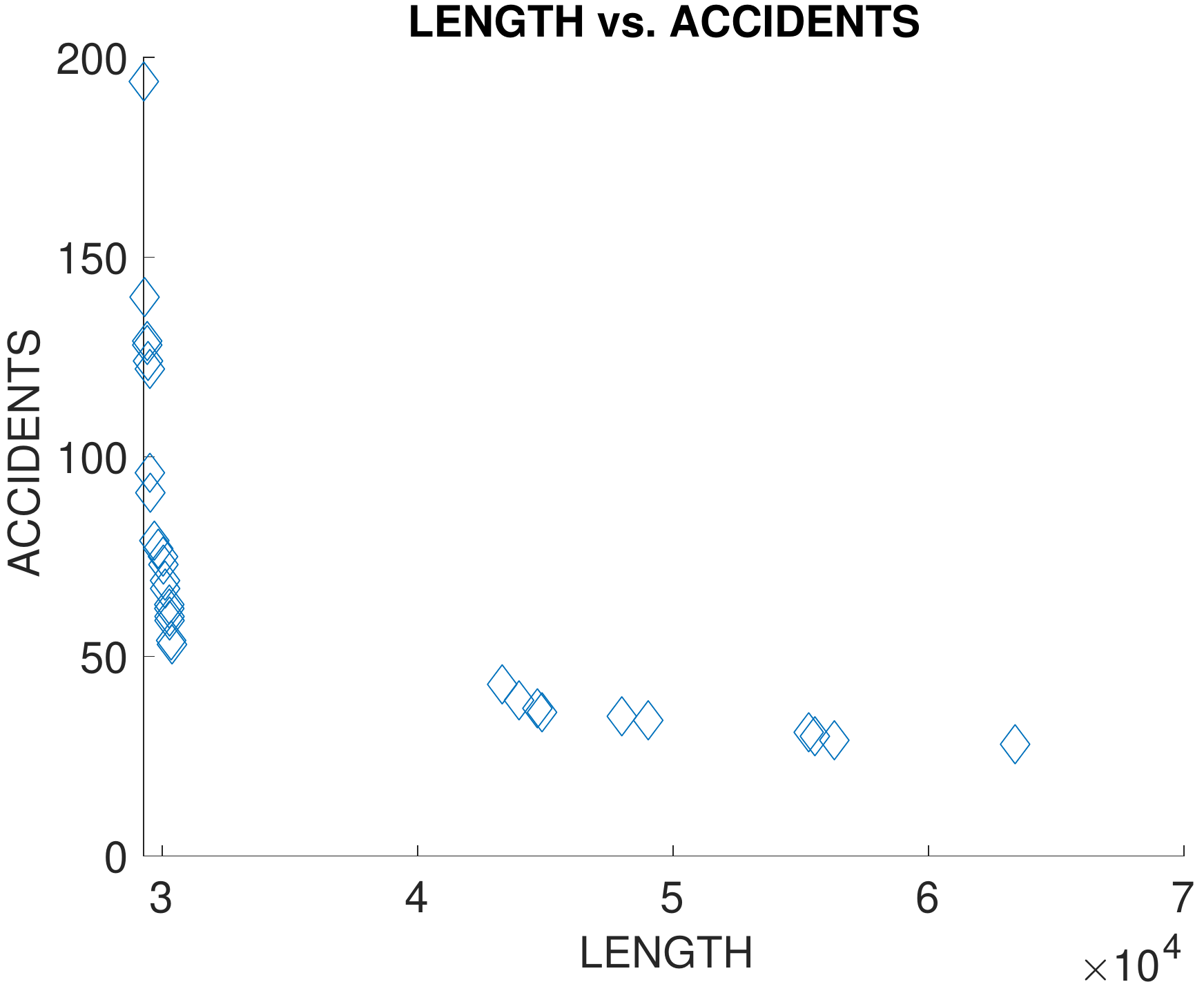}
}\qquad
\subfloat[Length vs. Accidents]
{
\label{fig:result:trueref:LEAC}
\includegraphics[width=0.2\textwidth]{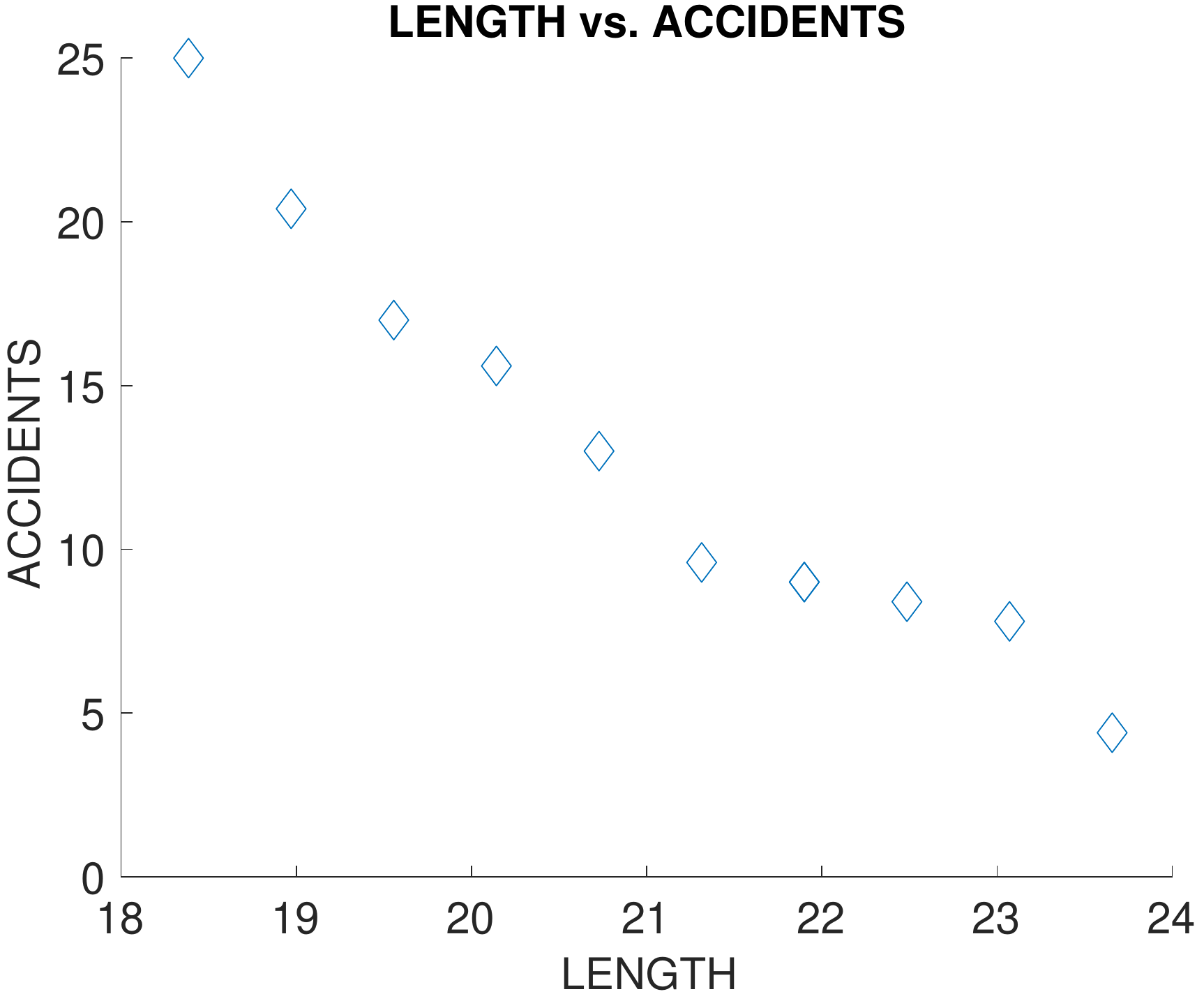}
}%
    \caption{Plots of the pairwise front between the Length-objective and the Accidents-objective on the real-map instance and the benchmarking instance: ASLETISMAC$\_$NO$\_$X14$\_$Y14$\_$PM$\_$K3$\_$BF}
        \label{fig:results:comparison:LEAC}
\end{figure}




\section{Conclusion}
\label{sec:conclusion}
In this paper, we present a variable many-objective pathfinding benchmark problem framework together with a model which can be easily transferred to a real-world related navigation problem on actual map data. The benchmark is scalable and can be used to analyse many-objective optimisation techniques for route planning and navigation. Different obstacle types, as well as elevation functions, neighbourhoods and backtracking properties, can be adjusted according to the needed complexity. We proposed five objective functions for the benchmark related to real-world goals when planning a route. Furthermore, we obtained the true Pareto-fronts for several benchmark instances which we also provide in the supplementary material. 

Additionally, we applied three evolutionary algorithms to minimise five objectives and compared the results with the obtained true Pareto-front of several benchmark problem instances. Also, we transferred the benchmark's characteristics to real-world data by adding further information to an obtained OpenStreetMap data graph. We also applied the algorithms with the same parameters and could obtain promising results.\\
In the future, we will work on advanced algorithms and operators to work on our benchmark. Besides, we aim to analyse more extensive and more complex instances of the problem, specifically the instances with enabled backtracking. As this will increase the search space by several magnitudes, we also aim to investigate more sophisticated methods to obtain the true Pareto-fronts of more complex instances. Additionally, we want to increase the number of objectives and will also investigate dynamically changing objectives, e.g. traffic data, which is often taken into account when using current route planning systems. Furthermore, we will work on robustness measurements when the environment contains uncertainties. We also plan to analyse the real-world data further to provide more realistic benchmark instances. Eventually, we want to encourage the community to analyse the instances even further and apply their algorithms on the problem.\\




%

\appendices
\section{}


\ifCLASSOPTIONcaptionsoff
  \newpage
\fi



\bibliographystyle{IEEEtran}
\bibliography{references}
%



%

\begin{IEEEbiography}[{\includegraphics[width=1in,height=1.25in,clip,keepaspectratio]{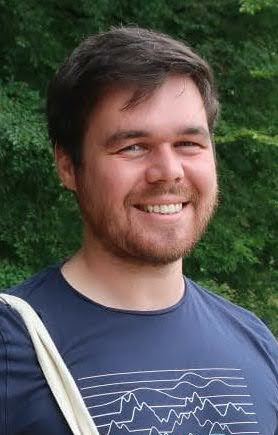}}]{Jens Weise}
received his B.Sc. degree in computer systems in engineering and his M.Sc. in medical systems engineering at the Otto von Guericke University Magdeburg, in 2013 and 2017 respectively.
His current research includes many-objective pathfinding and route planning using evolutionary multi-objective optimisation methods. He also researches on methods of self-organisation in the field of factory planning.
\end{IEEEbiography}

\begin{IEEEbiography}[{\includegraphics[width=1in,height=1.25in,clip,keepaspectratio]{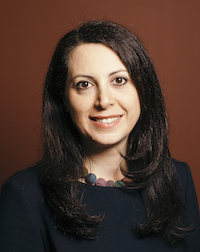}}]{Sanaz Mostaghim}
is a full professor of computer science at the Otto von Guericke University Magdeburg, Germany. She holds a PhD degree in electrical engineering and computer science from the University of Paderborn, Germany. She worked as a postdoctoral fellow at ETH Zurich in Switzerland and as a lecturer at Karlsruhe Institute of technology (KIT), Germany. Her research interests are in the area of evolutionary multi-objective optimization, swarm intelligence,
and their applications in robotics, science and industry. She serves as an associate editor for the IEEE Transactions on Evolutionary Computation, IEEE Transactions on Cybernetics, IEEE Transactions on System, Man and Cybernetics: Systems and IEEE Transactions on Emerging Topics in Computational Intelligence.
\end{IEEEbiography}







\end{document}